\documentclass[11pt,a4paper]{article}
\usepackage[hyperref]{naaclhlt2019}
\usepackage[utf8]{inputenc}
\usepackage[T1]{fontenc}
\usepackage{times}
\usepackage{latexsym}
\usepackage{xargs} 
\usepackage[colorinlistoftodos,prependcaption,textsize=small]{todonotes}
\usepackage{url}
\usepackage{amsmath}
\usepackage{algorithm}
\usepackage{algorithmic}
\usepackage{mathtools}
\usepackage{enumitem}
\usepackage{breqn}

\DeclarePairedDelimiter\floor{\lfloor}{\rfloor}

\title{Misspelling Oblivious Word Embeddings}
\vspace{1.5cm}
\author{
   Bora Edizel\\
   Facebook AI, London\\
   {\tt edizel@fb.edu} \\
   \And
   Aleksandra Piktus\\
   Facebook AI, London \\
   {\tt piktus@fb.com} \\ 
   \And
   Piotr Bojanowski\\
   Facebook AI, Paris\\
   {\tt bojanowski@fb.com} \\ 
   \AND
   Rui Ferreira\thanks{\hspace{1em}This work was carried out when the author was working as an employee at Facebook London.} \\
   Spotify, London\\
   {\tt ruif@spotify.com}\\ 
   \And
   Edouard Grave \\
   Facebook AI, Paris \\
   {\tt egrave@fb.com} 
   \And
   Fabrizio Silvestri\\
   Facebook AI, London \\
   {\tt fsilvestri@fb.com} \\
}

\date{}
\aclfinalcopy
\def\aclpaperid{1348} %  Enter the acl Paper ID here
\begin{document}
\maketitle
\vspace{2cm}

\begin{abstract}
In this paper we present a method to learn word embeddings that are resilient to misspellings. Existing word embeddings have limited applicability to malformed texts, which contain a non-negligible amount of out-of-vocabulary words. We propose a method combining FastText with subwords and a supervised task of learning misspelling patterns. In our method, misspellings of each word are embedded close to their correct variants. We train these embeddings on a new dataset we are releasing publicly. Finally, we experimentally show the advantages of this approach on both intrinsic and extrinsic NLP tasks using public test sets.

%Existing word embeddings have limited applicability to malformed texts, which contain a non-negligible amount of out-of-vocabulary words. In this paper we present a method that leverages the popular FastText model to learn word embeddings resilient to misspellings. We propose to extend FastText with a supervised component, which embeds misspellings close to their correct variants.
%We train our embeddings on a new dataset we are releasing publicly. Finally, we experimentally show the advantages of this approach on both intrinsic and extrinsic NLP tasks using public test sets.
\end{abstract}

\section{Introduction}
\label{sec:intro}
Word embeddings constitute a building block of many practical applications across NLP and related disciplines. Techniques such as Word2Vec \cite{efficientMikolov2013,mikolov2013distributed} and GloVe~\cite{pennington2014glove} have been extensively used in practice. One of their drawbacks, however, is that they cannot provide embeddings for words that have not been observed at training time, i.e. \textit{Out-Of-Vocabulary} (OOV) words. In real-world tasks, the input text is often generated by people and misspellings, a common source of OOV words, are frequent (e.g. \cite{cuerzan2004spelling} report that misspellings appear in up to 15\% of web search queries). As a consequence, the quality of downstream applications of word embeddings in real-world scenarios diminishes.

Simply allowing the inclusion of misspellings into corpora and vocabularies in existing methodologies might not provide satisfactory results. The sparsity of misspellings would most likely prevent their embeddings from demonstrating any interesting properties. Trying to balance the representation of misspellings with the representation of correctly spelled variants in training data by artificially introducing misspelled variants for every word in the corpus would on the other hand cause up to an exponential growth in the size of the training data, making training of the models infeasible.

% Similar motivations - leveraging intrinsic similarities of certain groups of words without blowing up the size of the model and training time, inspired the development of FastText. There, the morphology of the words is leveraged in order to build vector representations sensitive to 
% %aware of
% the subword structure. 
% % A hash-based trick is applied to store the generated subwords, preventing an uncontrolled increase in the vocabulary size. \todo{BE: I think we don't need this sentence here.}
% Subwords can than be 
% % re-composed 
% used 
% to generate representations of words unobserved at training time. As a result the probability of incurring an OOV word is close to $0$ (see Section \ref{sec:fasttext} for a detailed explanation of FastText embeddings). This approach constitutes a good baseline for the problem of misspellings since for low edit distance pairs we can expect the majority of subwords of a misspelling and a correction respectively to be the same and as a consequence - their vector representations to be close in the vector space.

% Although FastText can be considered a good baseline for the problem of embedding misspelled words, our intuition is that it will perform well only for the cases where dominant morphemes remain unchanged between misspellings and their corrections. The method may not be sufficient for high edit-distance, or intra-morpheme misspellings.

To address this deficiency, we propose Misspelling Oblivious (word) Embeddings (\textbf{MOE}), a new model combining FastText~\cite{bojanowski2017enriching} with a supervised task which embeds misspellings close to their correct variants. We carry out experiments on well established tasks and on their variants adapted to the misspellings problem. We also propose new methods of evaluating embeddings specifically designed to capture their quality on misspelled words. We train \textbf{MOE} embeddings on a new dataset we are releasing publicly. Finally, we experimentally show the advantages of this approach on both intrinsic and extrinsic NLP tasks using public test sets. Summarizing, we propose the following contributions:
\begin{itemize}[noitemsep]
\item a novel problem and a non-trivial solution to building word embeddings resistant to misspellings;
% and is able to model misspellings and embed them in the same vector space as correctly spelled words;
\item a novel evaluation method specifically suitable for evaluating the effectiveness of \textbf{MOE};
% consisting in measuring how high in the ranking of the nearest neighbors for a misspelling is positioned the correct word.
\item a dataset of misspellings \footnote{\url{https://bitbucket.org/bedizel/moe}} to train \textbf{MOE}.
\end{itemize}

The reminder of this paper is structured as follows. Section \ref{sec:related} gives an overview of the word embeddings literature. In Section \ref{sec:fasttext} we introduce Word2Vec and FastText models. We introduce the \textbf{MOE} model in Section \ref{sec:mowe}. Section \ref{sec:data} contains the descriptions of datasets we trained on and section \ref{sec:results} contains the description of experiments we conducted and their results. In Section \ref{sec:conclusions} we present our conclusions and plans for further research.

\section{Related Work}
\label{sec:related}
One of the first works to introduce the concept of a distributed representation for symbolic data was~\cite{hinton1986learning}. 
Later, the Information Retrieval community proposed techniques of embedding words into a vector space. Latent Semantic Indexing \cite{deerwester1990indexing} was one of the most influential works in this area. \cite{bengio2003neural} introduced the first neural language model which jointly learned word embeddings. Although such a language model was outperforming the baselines, it was not practical because of long training time requirements. \cite{collobert2008unified} proposed new neural architectures for word embeddings and showed that pre-trained word embeddings can be very valuable for some downstream NLP tasks. Word2Vec~\cite{mikolov2013distributed,efficientMikolov2013} got very popular both because of its effectiveness and its ability to train models on very large text corpora efficiently.
\cite{levy2014neural} showed that Word2Vec's skip-gram with negative sampling model (SGNS) is implicitly equivalent to word co-occurrence matrix factorization.
Besides neural approaches, \cite{pennington2014glove} proposed an SVD based architecture which gained a lot of attention because it allowed to effectively consider the popularity of each word in the model definition.
% Some of the recent language models e.g. \cite{peters2017semi,DBLP:journals/corr/JozefowiczVSSW16} make use of pre-trained word embeddings in their architectures.
% \cite{le2014distributed} proposed a model for embedding paragraphs and documents, \cite{cho2014learning} proposed a model for embedding phrases. Moreover, another extension of Word2Vec \cite{DBLP:conf/sigir/GrbovicDRSBFOYO16} proposed a model which embeds ads and queries into the same vector space.

FastText \cite{bojanowski2017enriching} is a popular, recent proposal in the area of word embeddings. FastText introduces subword-level features to the Word2Vec framework which enables building embeddings for OOV words (see details in Section~\ref{sec:fasttext}). An alternative approach, also capable of yielding representations for OOV words, is MIMIK~\cite{pinter2017mimicking}. MIMICK learns a function from input words to embeddings by minimizing the distance between embeddings produced by a char-based approach and the pre-trained embeddings. As opposed to \textbf{MOE}, MIMICK does not support misspellings explicitly and it requires a set of pre-trained embeddings as input. We consider MIMICK to be a viable alternative to FastText which deserves future work exploring its performance on misspelled text.

\section{Misspelling Oblivious Embeddings}
\subsection{The FastText Model}
\label{sec:fasttext}
Our current work can be viewed as a generalization of FastText, which, in turn, extends the the skip-gram with negative sampling (SGNS) architecture, proposed as a part of the Word2Vec framework. In this section we will briefly discuss major additions to SGNS introduded by FastText.

Let $V$ be a vocabulary of words and $T = w_1, w_2, \ldots, w_{|T|}$ be a text corpus, represented as a sequence of words from $V$. We define the context of a word $w_i\in V$ as $C_i = \{ w_{i - l}, \ldots, w_{i - 1}, w_{i + 1}, \ldots, w_{i + l}\}$ for some $l$ set as a hyperparameter. In the SGNS model, a word $w_i$ is represented by a single embedding vector $\mathbf{v}_i$ equivalent to the input vector of a simple feed-forward neural network, trained by optimizing the following loss function:

\begin{equation}
\begin{split}
\label{loss:word2vec}
L_{W2V} := \sum\limits_{i = 1}^{|T|}  \sum\limits_{w_{c} \in C_i} [ \ell(s(w_i, w_{c})) + \\
\sum\limits_{w_{n} \in N_{i,c}} \ell(-s(w_i, w_{n})) ]
\end{split}
\end{equation}
where $\ell$ denotes the logistic loss function $\ell(x) = log(1 + e^{-x})$ and $N_{i,c}$ is a set of negative samples drawn for the current word $w_i$ and its context $w_c \in C_i$. $s$ is the scoring function, which for SGNS is defined as the the dot product $\mathbf{v}_{i}^T \mathbf{u}_c$, where $\mathbf{u}_{c}$ is an output vector associated with the word $w_c$ and $\mathbf{v}_i$ is an input vector associated with the word $w_i$. Therefore, $s\left(w_i, w_{c}\right)=\mathbf{v}_{i}^T \mathbf{u}_c$.

In FastText, we additionally embed subwords (also referred to as character $n$-grams) and use them to construct the final representation of $w_i$. Formally, given hyperparameters $m$ and $M$ denoting a minimum and a maximum length of an $n$-gram respectively, the FastText model embeds all possible character $n$-grams of the word such that $m \le n \le M$. E.g. given $m = 3$, $M = 5$ and the word  \emph{banana}, the set of $n$-grams we consider is $ban, ana, nan, bana, anan, nana, banan, anana$. Now, let $\mathcal{G}_{w_i}$ denote the set of all $n$-grams of a word $w_i \in V$ plus the word itself (e.g. $\mathcal{G}_{\mathrm{banana}}$ is the set defined in the example above plus the word \emph{banana} itself). Given $\mathcal{G}_{w_{i}}$, FastText scoring function for a word $w_i$ and a context $w_c$ is defined as follows:

\begin{equation}
\label{scoring:fasttext}
s_{FT}(w_{i}, w_{c}) := \sum\limits_{\mathbf{v}_g, g \in \mathcal{G}_{w_i}}{\mathbf{v}_g}^{T} \mathbf{u}_c
\end{equation}

\noindent Therefore, the representation of $w_i$ is expressed through the sum of the representations of each of the $n$-grams derived from $w_i$ plus the representation of $w_i$ itself. FastText optimizes the loss function in Eq.\ref{loss:word2vec}, but uses the scoring function $s_{FT}$ defined in Eq.\ref{scoring:fasttext}. An extensive experimentation has shown that FastText improves over the original Word2Vec skip-gram model. The loss function of FastText will be referred to as $L_{FT}$ throughout the rest of this work.

\subsection{The MOE Model}
\label{sec:mowe}
As was shown empirically in the FastText paper, the $n$-grams which impact the final representation of a word the most in FastText correspond to morphemes. Based on this observation, we hypothesize that although FastText can capture morphological aspects of text, it may not be particularly resistant to misspellings which can occur also withing the dominant morphemes. In this section, we present the architecture of our model - \textbf{MOE} or Misspelling Oblivious (word) Embeddings. \textbf{MOE} holds the fundamental properties of FastText and Word2Vec while giving explicit importance to misspelled words.

\paragraph{Loss Function.}
The loss function of \textbf{MOE} is a weighted sum of two loss functions: $L_{FT}$ and $L_{SC}$. $L_{FT}$ is the loss function of FastText which captures semantic relationships between words. $L_{SC}$ or the spell correction loss aims to map embeddings of misspelled words close to the embeddings of their correctly spelled variants in the vector space. We define $L_{SC}$ as follows:

\begin{equation}
\begin{split}
\label{loss:spellCorrection}
L_{SC} :=  \sum_{(w_m, w_e) \in M} [ \ell(\hat{s}(w_m, w_e)) \ + \\
\sum_{w_{n} \in N_{m,e}} \ell(-\hat{s}(w_m, w_{n})) ]
\end{split}
\end{equation}

\noindent where $M$ is a set of pairs of words $(w_m, w_e)$ such that $w_e \in V$ is the expected (correctly spelled) word and $w_m$ is its misspelling. $N_{m,e}$ is a set of random negative samples from $V\setminus\{w_m, w_e\}$. $L_{SC}$ makes use of the logistic function  $\ell(x) = log(1 + e^{-x})$ introduced in Section \ref{sec:fasttext}. The scoring function $\hat{s}$ is defined as follows:

\begin{equation}
\label{scoring:moe}
\hat{s}(w_{m}, w_{e}) = \sum\limits_{\mathbf{v}_g, g \in \mathcal{\hat{G}}_{w_m}}{\mathbf{v}_g}^{T} \mathbf{v}_e 
\end{equation}
% \todo[color=blue!20]{EG: Does the vector $v_e$ uses subwords?}
% \todo[color=green!20]{AP: nope,  we just take subwords of the misspellings}

\noindent 
% It is worth highlighting that unlike in the FastText scoring function (eq.\ref{scoring:fasttext}), in the above formula both vectors representing members of $\mathcal{G}_{w_m}$ and the vector representing the correction $w_e$ come from the input matrix. What should be also clear from the above formulation is that we don't actually attempt to embed the misspellings themselves - we only embed their subwords. \todo{BE: Does not sound correct to me, will correct soon!}\todo[color=green!20]{AP: I wrote it (that was my impression looking at the code) but feel free to remove this line if incorrect}
where $\mathcal{\hat{G}}_{w_m} :=  \mathcal{G}_{w_m} \setminus \{w_m\}$. Therefore, the scoring function is defined as the dot product between the sum of input vectors of the subwords of $w_m$ and the input vector of $w_e$. Formally, the term $\ell(\hat{s}(w_m, w_e))$ enforces predictability of $w_e$ given $w_m$. Intuitively, optimizing $L_{SC}$ pushes the representation of a misspelling $w_m$ closer to the representation of the expected word $w_e$. It is also worth mentioning that embeddings for $w_m$ and $w_e$ share the same parameters set. The complete loss function of \textbf{MOE}, $L_{MOE}$, is defined as follows:

\begin{equation}
\label{loss:move}
L_{MOE} := (1 - \alpha) L_{FT} + \alpha \frac{|T|}{|M|} L_{SC} 
\end{equation}

Optimizing the loss functions $L_{FT}$ and $L_{SC}$ concurrently is not a straightforward task. This is because two different loss functions iterate over two different datasets: the text corpus $T$, and the misspellings dataset $M$. 
The optimization process should be agnostic to the sizes of $T$ and $M$ in order to prevent results from being severely affected by those sizes. Therefore, we scale $L_{SC}$ with the coefficient $|T|/|M|$. 
This way the importance of a single Stochastic Gradient Descent (SGD) update for $L_{FT}$ becomes equivalent to a single SGD update for $L_{SC}$. Moreover, $\alpha$ is the hyperparameter which sets the importance of the spell correction loss $L_{SC}$ with respect to $L_{FT}$ thus making \textbf{MOE} a generalization of FastText. 

% Each iteration, we make an SGD update to  $L_{W2V}$ with coefficient $\alpha$ while making an update for $L_{W2V}$ after each $|T|/|M|$ iterations.
% \subsection{Optimization}
% We optimize the loss by following the same technique presented in the original FastText paper \cite{bojanowski2017enriching}. 
% \improvement[inline]{Make sure that in the scoring function for the misspelling we clearly point out that both original and misspelling embeddings are coming from the same matrix.} -- Done, Bora.
% \begin{algorithm}
% \caption{MOE Scheduler}          
% \label{alg:scheduler} 
% \begin{algorithmic}
% \STATE $counter \gets 0$
% \STATE $notConverged \gets True$
% \WHILE{$notConverged$}
% \STATE $notConverged \gets SGD(L_{W2V}, \alpha)$
% \IF{counter \% $\frac{|T|}{|M|} = 0$} 
% 	\STATE $SGD(L_{SC}, 1-\alpha)$
% \ENDIF
% \STATE $counter \gets counter + 1$
% \ENDWHILE
% \end{algorithmic}
% \end{algorithm}
% Scheduler Algorithm\ref{alg:scheduler} optimizes $L_{MOE}$ using Stochastic Gradient Descent (SGD).

%\section{Experimental Setup}

\section{Data}
\label{sec:data}
As mentioned in Section \ref{sec:mowe}, \textbf{MOE} jointly optimizes two loss functions, each of which iterates over a separate dataset - a corpus of text for the FastText loss $L_{FT}$ and a set of pairs (misspelling, correction) for the spell correction loss $L_{SC}$. In this section, we will briefly discuss how we obtain each of these datasets.

\subsection{English text corpus}
We use an English Wikipedia dump\footnote{\url{dumps.wikimedia.org}} as the text corpus $T$ to optimize $L_{FT}$. The baseline FastText model is also trained on this dataset. Matt Mahoney's perl script\footnote{\url{http://mattmahoney.net/dc/textdata}} is used for pre-processing the raw Wikipedia dump. After pre-processing, the training corpus consists of $|T|=4,341,233,424$ words. When generating the vocabulary $V$ based on the corpus, we apply a frequency threshold of $5$.  After deduplication and thresholding, the size of the vocabulary for our corpus is $|V| = 2,746,061$ words. We also apply progressive subsampling of frequent words in order to not assign too much importance to highly frequent words. %It is worth mentioning that in FastText, subwords are not considered to be members of $V$. Similarly, in \textbf{MOE} we don't consider the subwords of misspellings as members of $V$. \todo{BE: Do we need to mention about last two sentences?}

\subsection{Misspelled data generation}
The misspellings dataset $M$ consists of a set of pairs $(w_m, w_e)$, where $w_e \in V$ represents a (presumably correctly spelled) word from the vocabulary and $w_m$ is a misspelling of $w_e$. Given the size of $V$, we opt for generating misspellings in an automated fashion, using an in-house misspellings generation script. The script is based on a simple error model, which captures the probability of typing a character $p_m$ when a character $p_e$ is expected (note that it's possible to have $p_m == p_e$), given a context of previously typed characters. The script is capable of generating misspellings of targeted edit distance for an input word $w_i$. In the reminder of this section, we'll discuss details of the script implementation.

\paragraph{Error model.} In order to create the error model, we first mine query logs of a popular search engine\footnote{\url{https://www.facebook.com}} and identify cases where a query was manually corrected by a searcher. We then pivot on the modified character and for each such modification we save a triplet $(c, p_m, p_e)$, where $p_m$ is the pivot character before modification, $p_e$ is the the target character after modification and $c$ represents up to 3 characters preceding the pivot in the original query. E.g. given a query \textit{hello worjd} corrected to \textit{hello world}, we would generate four triplets: $[(wor, j, l),(or, j, l),(r, j, l), (\epsilon, j, l))]$, where $\epsilon$ represents an empty word. Similarly, we create triplets by pivoting on characters which have not been modified. After processing all available logs, we count each unique triplet. For each unique pair $(c, p_m)$ of a context and a pivot, we then create a \emph{target list} consisting of all possible targets $p_e$, each associated with its probability calculated based on counts. We then sort each target list in the order of decreasing probability.

\paragraph{Injecting misspellings.} Let's consider a word $w_i \in V$ that we want to misspell. For each character $p\in w_i$, we take it's longest possible context $c$ (up to 3 characters) and we look up the \emph{target list} corresponding to $(c, p)$. We then proceed along the \emph{target list}, summing up the probabilities of subsequent targets until the sum is greater or equal to a randomly selected target probability $tp \in [0.0, 1.0]$. We then choose the corresponding target $t$ as a replacement for $p$ (note that in the majority of the cases $t == p$). We repeat this process for every word from $V$.

In order to respect real distributions of words in the text corpus $T$, we set the number of misspellings generated for each word $w_i \in V$ to be equal to the square root of the number of appearances of $w_i$ in $T$. The total size of misspellings dataset generated in this fashion is $|M|=20,068,964$ pairs. We make the dataset of misspellings publicly available at \url{https://bitbucket.org/bedizel/moe}.

\section{Experiments}
\label{sec:results}
In this section, we describe the experimental set up used for training our models and the experiments we conducted.

\subsection{Experimental setup}
We use FastText\footnote{\url{https://fasttext.cc/}} as a baseline for comparison since it can generate embeddings for OOV words which makes it potentially suitable for dealing with misspellings.
We train the baseline model using the default hyperparameters provided by the authors. We consider character $n$-grams of lengths between $m = 3$ and $M=6$, and we use $5$ negative samples for each positive sample. Training \textbf{MOE} requires optimizing two loss functions $L_{FT}$ and $L_{SC}$ jointly. For optimizing $L_{FT}$, we use the same parameters as in the baseline. Additionally, to optimize $L_{SC}$, we experiment with $5$ negative samples per positive sample. We sweep over a range of values for the coefficient combining the two losses: $\alpha\in\{0.01, 0.05, 0.1, 0.5, 0.25, 0.5, 0.75, 0.95, 0.99\}$. Both FastText and \textbf{MOE} are trained using Stochastic Gradient Descent with a linearly decaying learning rate for $5$ epochs to learn vectors with $300$ dimensions. We evaluate the performance of \textbf{MOE} on the following tasks: (intrinsic) Word Similarity, Word Analogy and Neighborhood Validity; (extrinsic) POS Tagging of English sentences.

We report the overlap between the misspellings seen at training time and misspellings present in tests in Table~\ref{table:overlap}.

\begin{table}[t]
\centering
\begin{tabular}{c|c}
Test set & \% of unseen \\ \hline
WS353 $r = 0.125$ & 25.05 \\
WS353 $r = 0.250$ & 57.06 \\
WS353 $r = 0.375$ & 64.37 \\
RareWord $r = 0.125$ & 44.67 \\
RareWord $r = 0.250$ & 70.18 \\
RareWord $r = 0.375$ & 78.84 \\
Word Analogies & 50.71 \\
Neighborhood Similarity & 69.1 \\
\end{tabular}

\caption{Percentages of test misspellings unobserved at training time per test set. The $r$ parameter indicates variants of respective word similarity test sets.}
\label{table:overlap}
\end{table}

\subsection{Intrisic Evaluation}
We evaluate MOE on two classic intrinsic tasks, namely Word Similarity and Word Analogy and on a novel intrinsic task evaluating the distance between vector embeddings of misspellings and their correctly spelled variants.

\paragraph{Word Similarity.}
In the word similarity task, we evaluate how well embeddings generated by \textbf{MOE} can capture the semantic similarity between two words. For this purpose, we use two datasets: (i) \textit{WS353} \cite{finkelstein2001placing}, and (ii) \textit{Rare Words (RW)} \cite{luong2013better}. Both datasets contain pairs of words $w_a$, and $w_b$ annotated with a real value in the range $[0, 10]$ representing the degree of similarity between $w_a$ and $w_b$ as perceived by human judges.

\begin{figure}[tb!]
\centering
\includegraphics[width=1.0\columnwidth]{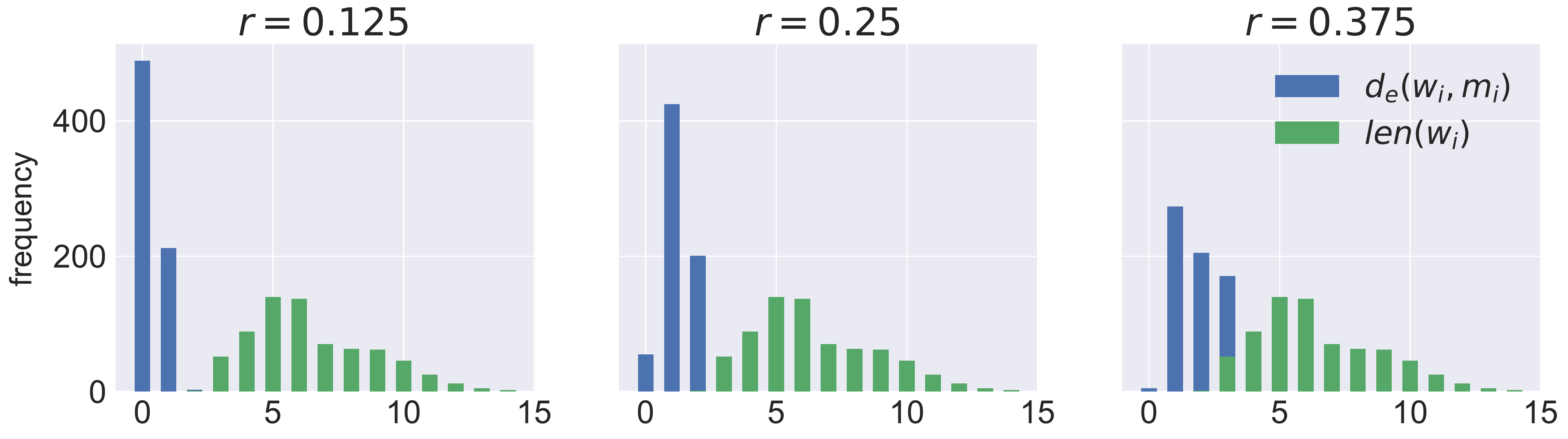}
\includegraphics[width=1.0\columnwidth]{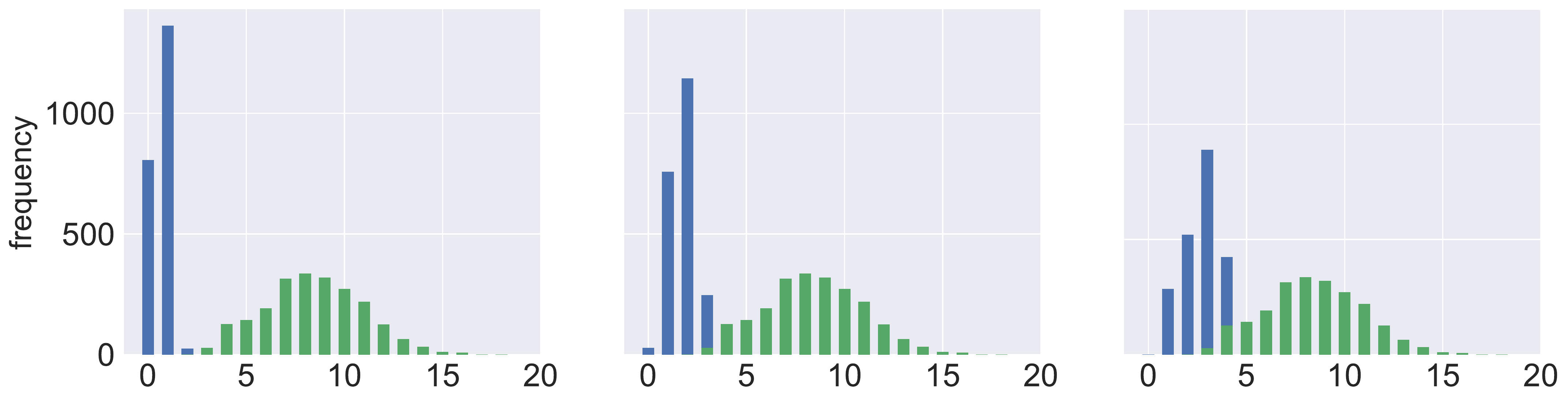}
\caption{Distribution of edit distances $d_e(w_i, m_i)$ and lenghts of words $len(w_i)$ for WS353 variants (Top) and RW variants (Bottom).}
\label{fig:datasetCharacteristics}
\end{figure}

In order to evaluate how resilient our method is to spelling errors, for each pair of words $(w_a, w_b)$ in the dataset, we provide a respective pair of misspellings $(m_a, m_b)$. The misspellings are mined from search query logs of a real-world online search service. When desired misspellings are not available in the logs, we synthetically generate them using the same script we used to generate the set $M$ (see Section \ref{sec:data} for details). We create 3 misspelled variants of both WS353 and RW datasets. In each variant we limit the ratio between the edit distance \cite{levenshtein1966binary} of the word and the misspelling $d_e(w_i, m_i)$ and the length of the word by a constant $r$, where $r\in \{0.125, 0.250, 0.375\}$, with $r=0$ representing the original dataset. More precisely for each $r$ we look for a misspelling which satisfies the following condition $d_e(w_i, m_i) = \floor*{r * len(w_i)}$. Effectively, if a word is too short to satisfy the condition, we preserve the original word (then $w_i$ = $m_i$). 
Histograms in Figure~\ref{fig:datasetCharacteristics} show the actual distribution of edit distances and lengths of words. 
As expected, edit distance increases steeply with the increase of $r$ value. Edit distances are higher for the RW dataset since in average the length of words in RW is higher than on average length of words in WS353. Also, we observe that for $r=0.125$, a significant portion of the words is not changed.

We conduct experiments for different values of the hyperparameter $\alpha$ which sets the trade-off between $L_{FT}$ and $L_{SC}$, i.e. the importance assigned to semantic loss and misspelling loss. In the experiments, results corresponding to $\alpha=0$ represents our baseline, FastText, since for $\alpha = 0$ the loss $L_{MOE}$ is equal to $L_{FT}$.

\begin{figure*}[tb!]
\centering
\includegraphics[width=\textwidth]{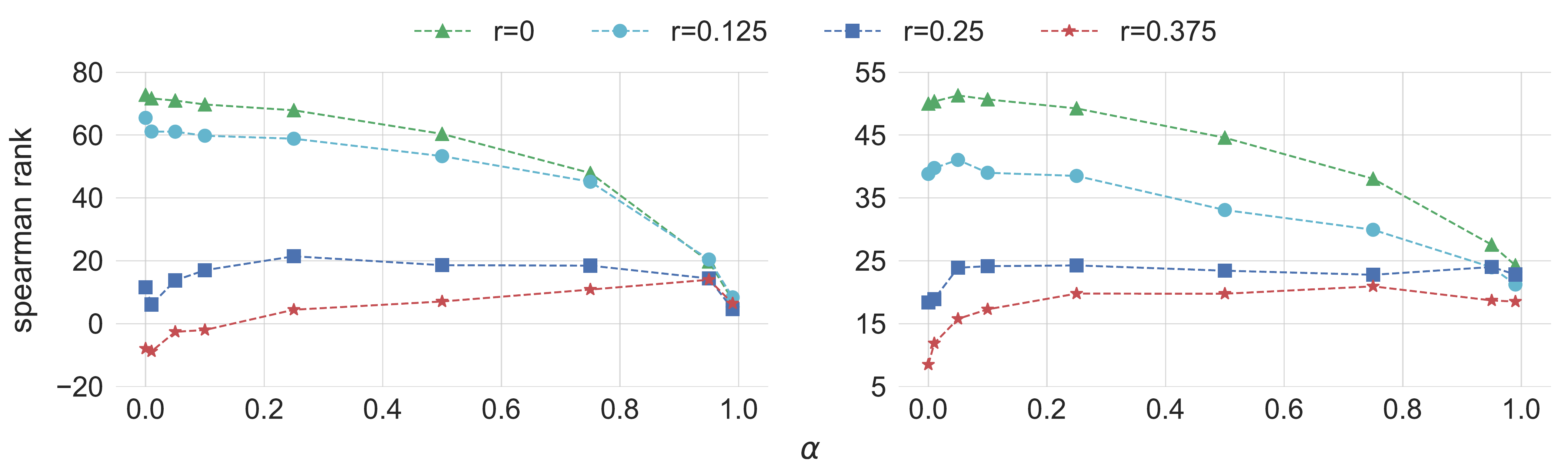}
\caption{Experimental results for word similarity task for WS353 (Left) and RW (Right). $\alpha=0$ values represent our baseline, FastText.}
\label{fig:ws353}
\end{figure*}

\begin{figure*}[tb!]
\centering
\includegraphics[width=\textwidth]{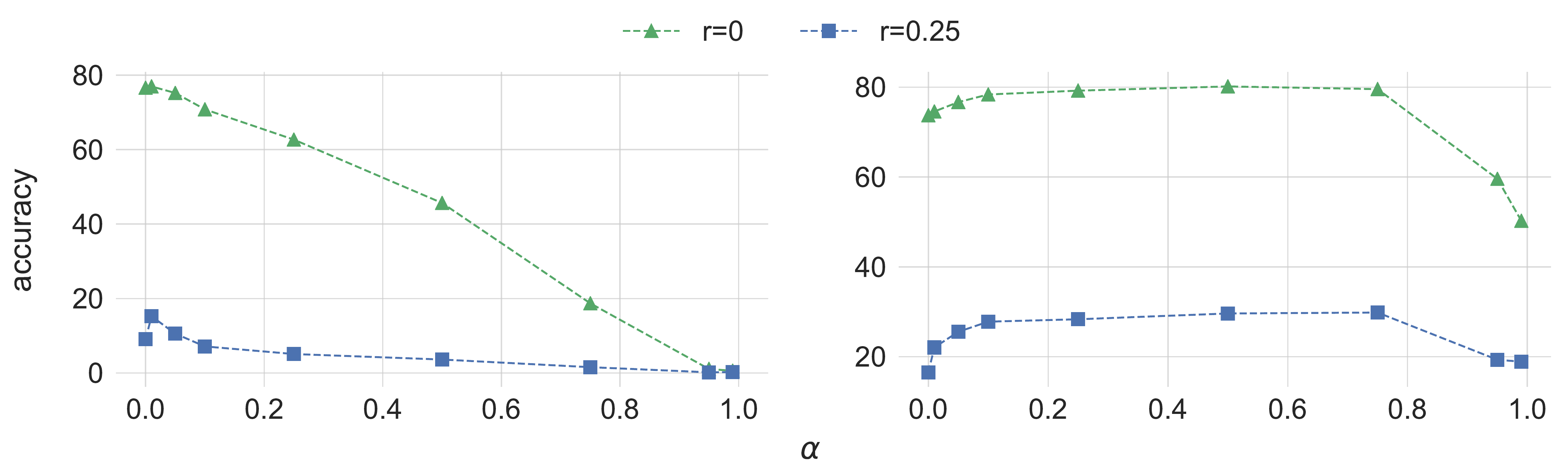}
% \caption{Experimental results for word analogy task. On the left side we present results for models with 5 negative samples for the MOE task ($N_{m,e}=5)$. On the right side $N_{m,e}=10$ respectively.}
\caption{Experimental results for word analogy task, Semantic (Left) and Syntactic (Right). $\alpha=0$ values represent our baseline, FastText.}
\label{fig:wordAnalogy-5-r0}
\end{figure*}

We measure the Spearman's rank correlation \cite{spearman1904proof} between the distance of an input pair of words and the human judgment score both for the original and the misspelled pair. Figure~\ref{fig:ws353} demonstrates the results of the word similarity task on the WS353 dataset.
We observe that \textbf{MOE} is improving over FastText for WS353 variants with $r=0.25$, and $r=0.375$, and degrading performance when $r=0$, and $r=0.125$, where the majority of the words is not changed (see Figure \ref{fig:datasetCharacteristics} for the edit distance distribution). As we expected, larger values of $\alpha$, corresponding to more attention given to misspellings during training, result in improvements for highly misspelled datasets.

For the RW dataset (Figure~\ref{fig:ws353}), we observe that for all the values of $r$, \textbf{MOE} improves over the FastText baseline when we set $\alpha=0.05$.
More specifically, when $r \in \{ 0, 0.125 \}$ and when $\alpha < \approx 0.1$, the proposed method improves over the baseline. When the amount of misspellings is higher, i.e., $r \in \{ 0.25, 0.375 \}$, \textbf{MOE} improves the results over the baseline for all of the $\alpha$ values. These results suggest that FastText may be a good baseline for dealing with low edit distance misspellings, however our model is better at capturing semantic relationships on higher edit distance misspellings. This is in line with our hypothesis presented in Section \ref{sec:mowe}.

\paragraph{Word Analogy.}

In addition to the word similarity, we also test the performance of \textbf{MOE} on the popular word analogy task introduced by \cite{efficientMikolov2013}. This task attempts to measure how good the embeddings model is at preserving relationships between words. 
A single test sample from the word analogy dataset consists of four words $A, B, C, D$, forming two pairs - $A, B$ and $C, D$, remaining in analogous relationships ("$A$ is to $B$ like $C$ is to $D$"). There are two types of relationships: (i) \textit{syntactic}, related to the structure of words; and (ii) \textit{semantic}, related to their meanings. \textit{banana, bananas, cat, cats} is an example of a syntactic test sample. In both pairs the fact that the second word is a plural version of the first constitutes a relationship between them. \textit{Athens, Greece, Berlin, Germany} is an example of a semantic test sample. The relationship which is being tested in this case is that between the capital of a country and the country itself.

In addition to analyzing the canonical variant of the word analogies test, we also introduce a modification which is suitable specifically to the misspellings use-case. Given a line $A, B, C, D$ from the original analogies dataset, we misspell the first pair of words, obtaining a line $A', B', C, D$, where $A'$ is a misspelling of $A$ and $B'$ is a misspelling of $B$. We want to test if the misspelled pair $A', B'$ preserves the relationship of the pair $C, D$. When generating misspellings we use a procedure similar to the one used for word similarities. We create one variant of the misspelled dataset, constraining the edit distance to $r=0.25$.

% \begin{figure}[tb!]
% \centering
% % \includegraphics[width=1.0\textwidth]{figures/wordAnalogy-5}
% \includegraphics[scale=0.4]{figures/wordAnalogy-5-r025}
% % \caption{Experimental results for word analogy task. On the left side we present results for models with 5 negative samples for the MOE task ($N_{m,e}=5)$. On the right side $N_{m,e}=10$ respectively.}
% \caption{Experimental results for word analogy task. $r = 0.25$}
% \label{fig:wordAnalogy-5-r025}
% \end{figure}

Experimental results for the canonical version of the word analogy task, presented in Figure~\ref{fig:wordAnalogy-5-r0}, show that \textbf{MOE} performs worse than FastText on the semantic analogy task. On the other hand, \textbf{MOE} performs better than the baseline on the syntactic analogies task. The results for the misspelled variant of the task show that,  
% are presented in Figure\ref{fig:wordAnalogy-5-r025}. 
the overall performance of both the baseline and \textbf{MOE} is worse than on the canonical variant. For low values of $\alpha \in \{0.01, 0.05\}$, \textbf{MOE} outperforms the baseline on the semantic task, achieving an over 67\% better score than FastText for $\alpha = 0.01$. \textbf{MOE} outperforms the baseline on the syntactic task for all tested values of $\alpha$, improving by over 80\% for $\alpha = 0.75$. For $\alpha = 0.01$, which achieved the best semantic result, the improvement on the syntactic task is over 33\%.\\[12pt]
\indent The trends that we observe both in the canonical and the misspelled variant of the word analogies task seem to validate our choice of the loss function for the \textbf{MOE} model. It is clear that the FastText component of the loss is indispensable to learn the semantic relationships between words. In fact, it is the only component of the loss function which attempts to learn these relationships. Therefore, decreasing it's importance (by increasing the value of $\alpha$) is reflected by a decay in the semantic analogies score. The spell-correction component of the loss function, on the other hand, leverages the relationship between correctly spelled words and their misspellings. As a side effect, it also adds additional subword information into the model. This explains our good performance on the syntactic analogies task. As our results on the misspelled variant of the task show, we improve over the baseline in understanding analogies on the misspelled words, which was one of the design principles for \textbf{MOE}.

\begin{table*}[t]
\centering
\begin{tabular}{c|c|c|c|c|c|c}
Test Data &\multicolumn{3}{c|}{100\% Misspelled} & \multicolumn{3}{c}{Original}  \\ \hline
Training Data & Original & 100\% Miss. & 10\% Miss. & Original & 100\% Miss. & 10\% Miss. \\ \hline
FastText, $\alpha=0.0$ & 30.47 & 79.71	&  65.70 &  94.33 & 57.16  & 94.14 \\
MOE, $\alpha=0.01$  & 29.04 & 80.66 & 67.94 & \textbf{94.55} & 59.11 & 94.21\\
MOE, $\alpha=0.05$ & 28.52 & \textbf{81.17} & \textbf{68.92} & 94.25 & 58.95 & 93.92 \\
MOE, $\alpha=0.1$  & 30.94 & 80.97 & 67.30 & 94.45 & 58.88 & \textbf{94.29} \\
MOE, $\alpha=0.25$ & 29.00 & 80.13 & 67.63 & 94.37 & 58.67 & 94.01 \\
MOE, $\alpha=0.5$  & 29.19 & 80.43 & 66.76 & 94.27 & 57.29 & 93.94 \\
MOE, $\alpha=0.75$ & 30.94 & 78.65 & 64.53 & 94.18 & 57.67 & 93.81 \\
MOE, $\alpha=0.95$ & 32.40 & 75.28 & 62.29 & 93.09 & 60.21 & 92.52 \\
MOE, $\alpha=0.99$ & \textbf{32.57} & 73.36 & 61.36 & 90.91	& \textbf{60.62} & 90.53\\ \hline
% diff & 2.1 & 1.28 & 2.06 & 0.22 & 3.46 & 0.15 \\
\end{tabular}
\caption{Performance on POS tagging task for UPOS tags using CRF. The models were trained on 100 epochs with an early stop (small difference on validation error) mechanism enabled. Considering F1 score, we evaluate on 2 variants of  test data: Original (correctly spelled) on the right hand side of the table and 100\% misspelled on the left hand side.}
\label{table:posTag}
\end{table*}

\paragraph{Neighborhood Validity.}
One of the explicit objectives of \textbf{MOE} is to embed misspellings close to their correct variants in the vector space. In order to validate this hypothesis, we check where in the neighborhood of a misspelling the correct word is situated. Formally, for a pair $(w_m, w_e)$ of a misspelling and its correction, we pick $k$ nearest neighbors of the misspelling $w_m$ in the embedding space using cosine similarity as a distance metric. We then evaluate the position of the correct word $w_e$ within the neighborhood of $w_m$ using two metrics:
\begin{itemize}[noitemsep]
  \item We use \textit{MRR} \cite{voorhees1999trec} to score the neighborhood of the embeddings of misspellings (we assign a score of $0$ if the correct word is not present).
  \item We also compute the \textit{neighborhood coverage} defined as the percentage of misspellings for which the neighborhood contains the correct version.
\end{itemize}

\begin{figure*}[tb!]
\centering
\includegraphics[width=\textwidth]{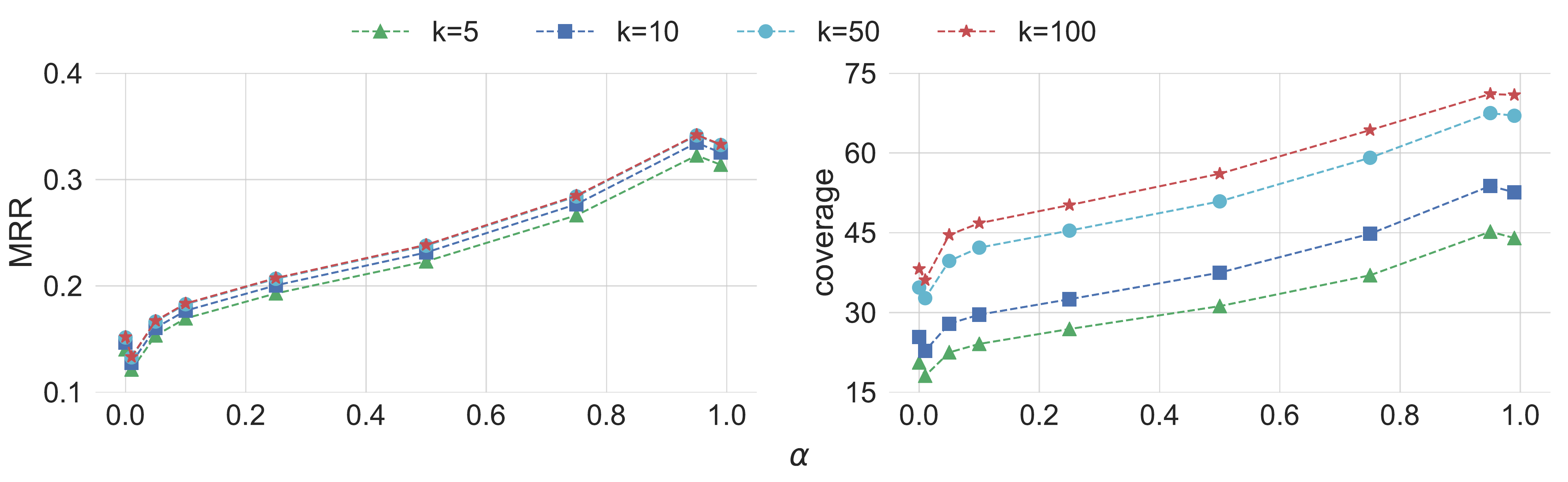}
\caption{Experimental results for the Neighborhood Validity task. $\alpha=0$ values represent our baseline, FastText. On the left hand side we present the resulting MRR scores. On the right hand size we present the results for the neighborhood coverage.}
\label{fig:kNeigh}
\end{figure*}

% We perform these steps both for the FastText baseline and for \textbf{MOE}. We test 5 variants of \textbf{MOE}, corresponding to $\alpha \in \{0.01,0.05,0.1,0.25,0.5\}$. All \textbf{MOE} models are trained with 5 negative samples for the misspellings task.
The test set contains $5,910$ pairs $(w_m, w_e)$ sampled from a collection of data coming from a real-world online service\footnote{\url{www.facebook.com}}. Figure~\ref{fig:kNeigh} shows experimental results for Neighbor Validity task.
We remind that $\alpha=0$ denotes the FastText baseline.

The test results confirm our hypothesis. We observe that MRR increases when more importance is given to the $L_{SC}$ component of the loss for any size of the neighborhood $k \in \{ 5,10,50,100 \}$. A similar trend can be observed for the neighborhood coverage task. We conclude that, on average, we're more likely to surface the correction using \textbf{MOE} than with FastText. What is more, whenever we are able to surface the correct version of a misspelled word, its position in the ranking is higher for \textbf{MOE} than for the FastText baseline.

\subsection{Extrinsic Evaluation}
\paragraph{POS Tagging.}
Finally, we evaluate \textbf{MOE} on a Part-of-Speech (POS) tagging task\footnote{\url{http://universaldependencies.org/conll17/data.html}}. To assess the impact of misspellings we artificially inject misspellings in the dataset. We train \textbf{MOE} on three different dataset variants: a non-misspelled dataset, to verify that \textbf{MOE} does not jeopardize the performance on correct words; a dataset where $10\%$ of words contain a misspelling, to simulate a realistic environment where some of the words are misspelled; and finally on a dataset where $100\%$ of words contain misspellings, to simulate a highly distorted environment. We use a state-of-the-art POS tagger \cite{ma2016end} consisting of a Conditional Random Fields (CRF) model where embeddings of the words in a sentence constitute observations and the tags to assign constitute the latent variables. This model adds a dependency on both layers of a Bi-LSTM component to the tag variables in the CRF. We evaluate the F1 score of the system for the three dataset variants we describe above. We test two different representations as input to the CRF: FastText (our baseline), and \textbf{MOE} embeddings. Our results are reported in Table~\ref{table:posTag}.

%We also trained a character-embeddings based version of the POS-tagger (as shown in~\cite{ma2016end}) observing trends consistent with those reported in the paper (we don't include these results due to lack of space). The complexity of these models, however, is much higher and training on realistic (large scale) datasets is not feasible in practice. FastText and \textbf{MOE}, instead, were fast both at training and inference time. This characteristics is particularly important as training on more data is always a good way of improving the performance of such systems and a more complex architecture does not easily support training on very large datasets.

We make the following observations based on the results of our experiments. Firstly, in the two extreme cases of the $100\%$ misspelled test and correct training and the correct test and $100\%$ misspelled training, \textbf{MOE} improves the F1 by $2$ and $3.5$ points respectively with respect to the FastText baseline. When the test data is $100\%$ misspelled, \textbf{MOE} always beats the baseline by up to $2.3$ points of F1. Also, in this case the loss in F1 with respect to the case where both the training and the test are correct is much less then when the training data does not contain misspellings. To be remarked is the F1 score difference in the more realistic case consisting of training data that is $10\%$ misspelled. In this case \textbf{MOE} attains a sensitive improvement of $2.3\%$ points of F1. Finally, \textbf{MOE} does not reduce the effectiveness of the CRF POS Tagger with respect to the FastText baseline when neither the training nor the test set are misspelled.
All in all, we have shown that \textbf{MOE} does not affect the effectiveness of the POS Tagger in the case of correctly misspelled words and improves sensitively the quality of the POS tagger on misspellings.

\section{Conclusion and Future Work}
\label{sec:conclusions}
One of the most urgent issues of word embeddings is that they are often unable to deal with malformed words which is a big limitation in the real-world applications. In this work, we proposed a novel model called \textbf{MOE}, which aims to solve a long-standing problem: generating high quality, semantically valid embeddings for misspellings.

In the experiments section, in the neighborhood validity task, we show that \textbf{MOE} maps embeddings of misspellings close to embedding of the corresponding correctly spelled word. Moreover, we show that \textbf{MOE} is performing significantly better than the FastText baseline for the word similarity task when misspellings are involved. For the canonical versions of the word similarity tasks, where misspellings are not involved, we show that \textbf{MOE} doesn't worsen the quality significantly for the WS353 dataset and improves over baseline for the RW dataset. In the word analogy task, \textbf{MOE} is able to preserve the quality of the semantic analogies similar to the baseline, while improving on the syntactic analogies. In the variant of the test where misspellings are involved, \textbf{MOE} outperforms the baseline on both semantic and syntactic questions. Finally, we have shown that \textbf{MOE} does not affect the effectiveness of the POS Tagger in the case of correctly spelled words and improves sensitively the quality of the POS tagger on misspellings.

In the future, we will test different ways of training embeddings for misspellings including the extension of the same technique to multi-lingual embeddings. We are going to test deep architectures to combine the $n$-grams in misspellings to better capture various interdependencies of $n$-grams and correct versions of words. Finally, we will assess the robustness of both character-based \cite{kim2016character} and context-dependent embeddings \cite{devlin2018bert}, \cite{Peters:2018} with respect to misspellings.

\end{document}